\pdfoutput=1
\documentclass[12pt,letterpaper]{article}

\usepackage[letterpaper,total={7in,10in}]{geometry}

\usepackage{graphicx}
\usepackage{helvet}
\usepackage{authblk}
\usepackage{hyperref}
\usepackage{amsmath}
\usepackage{amssymb}
\usepackage[super,comma,sort&compress]{natbib}
\usepackage{enumitem}

\hypersetup{
  hidelinks,
  pdftitle={An immune world model for multiscale forecasting and therapeutic hypothesis generation},
  pdfauthor={Taoyong Cui, Xi Wang, Zonghang Li, Jinchao Ding, Lingsen You, Yuzhi Xu, Wanghan Xu, Fang Wu, Kejun Ying, Wanli Ouyang, Pheng Ann Heng, Ling Yang, Zhenfei Yin, and Yingcheng Wu},
  pdfsubject={Immune world modeling and multiscale intervention forecasting},
  pdfkeywords={immune world model, AI Scientist, multiscale immune forecasting, tissue immune context, therapeutic hypothesis generation}
}
\makeatletter
\renewcommand{\maketitle}{\bgroup\setlength{\parindent}{0pt}
\begin{flushleft}
  \textbf{\@title}

  \@author
\end{flushleft}\egroup}
\makeatother

\title{An immune world model for multiscale forecasting and therapeutic hypothesis generation}
\date{}

\author[1,2,3,$\dagger$]{Taoyong Cui}
\author[4,$\dagger$]{Xi Wang}
\author[2]{Zonghang Li}
\author[1]{Jinchao Ding}
\author[10]{Lingsen You}
\author[4]{Yuzhi Xu}
\author[5]{Wanghan Xu}
\author[6]{Fang Wu}
\author[7]{Kejun Ying}
\author[3]{Wanli Ouyang}
\author[2,*]{Pheng Ann Heng}
\author[1,8,*]{Ling Yang}
\author[1,9,*]{Zhenfei Yin}
\author[1,10,*]{Yingcheng Wu}

\affil[1]{PhAI Labs, Inc., Palo Alto, CA, USA}
\affil[2]{Department of Computer Science and Engineering, The Chinese University of Hong Kong, Hong Kong SAR, China}
\affil[3]{Multimedia Laboratory, Department of Information Engineering, The Chinese University of Hong Kong, Hong Kong SAR, China}
\affil[4]{Department of Chemistry, New York University, New York, NY, USA}
\affil[5]{Shanghai Jiao Tong University, Shanghai, China}
\affil[6]{Department of Computer Science, Stanford University, Stanford, CA, USA}
\affil[7]{Harvard University, Cambridge, MA, USA}
\affil[8]{Princeton University, Princeton, NJ, USA}
\affil[9]{University of Oxford, Oxford, UK}
\affil[10]{Department of Pathology, Stanford University School of Medicine, Stanford, CA, USA}
\affil[$\dagger$]{These authors contributed equally}
\affil[*]{Correspondence: pheng@cse.cuhk.edu.hk (P.A.H.); yang@phai-labs.com (L.Y.); yin@phai-labs.com (Z.Y.); wuyc@phai-labs.com (Y.W.)}

\begin{document}
\maketitle

\section*{SUMMARY}

Immune therapies act across cell-intrinsic programs, tissue ecosystems, and patient-specific immune states, yet most predictors address these scales separately. We used a governed evolutionary AI Scientist to construct the Immune World Model, an action-conditioned model that learns how interventions move immune states across cellular, tissue, and individual levels. The Immune World Model--building Scientist searched candidate architectures and workflows, and the resulting world model was frozen before independent confirmation. The frozen model generalized to unseen interventions and biological contexts, recovered intervention-specific cellular programs, integrated cell and tissue information to improve ecosystem and patient-response prediction, and forecast unseen perturbation combinations. Immune World Model--guided analysis then combined measured perturbations with cross-axis inference to nominate IL-36$\gamma$ plus SIRP$\alpha$ inhibition as a complementary-axis therapeutic hypothesis, whereas a governed self-correction audit rejected every screened cytokine pair. The Immune World Model provides a framework for multiscale immune simulation that connects AI Scientist-driven model construction, intervention forecasting, and the generation of prospectively testable therapeutic hypotheses.

\section*{KEYWORDS}

immune world model, AI Scientist, multiscale immune forecasting, tissue immune context, therapeutic hypothesis generation

\section*{INTRODUCTION}

Therapeutic response in cancer emerges from interactions across cellular, tissue, and individual scales. Receptor availability and intracellular signaling determine how individual cells respond to an intervention; cellular composition, spatial organization, and intercellular communication shape the resulting tumor ecosystem; and host genetics, immune history, prior treatment, and disease state influence patient-level outcomes. Cancer immune phenotypes therefore reflect coordinated malignant, stromal, myeloid, and lymphoid programs.\cite{ref1,ref2,ref3} Response and resistance to checkpoint blockade likewise depend on immune exclusion, pretreatment T-cell state, genomic context, and the dynamics of tumor-reactive clones.\cite{ref4,ref5,ref6,ref7,ref8,ref9,ref10} Predictive models must therefore capture intervention-dependent state changes across this biological hierarchy.

Large single-cell atlases and perturbation screens provide the data needed to model these transitions. Geneformer, scGPT, scFoundation, Universal Cell Embeddings, and STACK learn transferable representations of cellular identity and regulatory state.\cite{ref11,ref12,ref13,ref14,ref15} Probabilistic integration and reference-mapping methods align batches, modalities, and atlas populations.\cite{ref16,ref17,ref18,ref19,ref20,ref21,ref22,ref23} scGen, compositional perturbation autoencoders, and GEARS predict cellular responses beyond observed training combinations from pooled genetic and chemical screens.\cite{ref24,ref25,ref26,ref27,ref28,ref29,ref30,ref31} These advances establish strong foundations for cellular representation and perturbation prediction, but a unified model must also connect cellular transitions to tissue ecology and individual response while distinguishing an intervention from the biological state on which it acts.

World models learn compact environmental representations and action-conditioned transition dynamics, enabling future states to be simulated in latent space.\cite{ref32,ref33} This formulation provides a natural framework for multiscale immune prediction. An immune world model can encode cellular, tissue, and individual context as a structured state, encode an intervention as a distinct action, learn the resulting transition, and decode the predicted state into gene programs, cellular composition, pathway activity, immune subtype, and treatment response. We refer to the propagation of intervention-conditioned information across these scales as ``World Model Rollout.'' Generalization can then be assessed by withholding actions, biological contexts, action--context pairs, or information from an entire scale. Constructing and evaluating such a system requires coordinated choices among representations, transition operators, training procedures, and evaluation criteria.

AI Scientist systems can organize evidence, select analytical tools, execute computational analyses, and evaluate intermediate results,\cite{ref34,ref35} and language-model agents have been applied to laboratory planning, automation, and prospective molecular design.\cite{ref36,ref37} We used Agent Genesis, a governed evolutionary AI Scientist, to construct the Immune World Model by evolving candidate Scientist workflows and immune-world-model configurations under versioned data manifests and prespecified evaluation gates. Here, Agent Genesis denotes the governed development framework, the Immune World Model denotes the learned biological model, and the Immune World Model Research Scientist denotes the fixed post-freeze workflow. The resulting Immune World Model configuration was frozen before independent confirmation and then evaluated on withheld interventions and contexts, cross-scale World Model Rollout, and action composition. Immune World Model--guided analysis further identified complementary immune axes and prioritized IL-36$\gamma$ plus SIRP$\alpha$ inhibition as a therapeutic hypothesis for prospective experimental evaluation.

\section*{RESULTS}

\subsection*{A structured immune world model links cellular, tissue, and individual states}

To connect intervention-dependent changes across cellular, tissue, and individual scales, we organized the Immune World Model around a structured biological state and an explicit intervention interface (Figures 1A and 1B). Agent Genesis evolved candidate Scientist workflows, and the retained Immune World Model--building Scientist compared model architectures before the model configuration was frozen. During AI4AI, candidates advanced from seed and tool-using agents to the Immune World Model--building Scientist; after model freezing, the fixed Immune World Model Research Scientist coordinated simulation, evaluation, execution repair, memory, and approved tools and prioritized selected hypotheses for prospective experimental testing. Model promotion and biological interpretation were performed under investigator oversight.

Within this architecture, multiscale immune data, perturbation and intervention data, patient-derived data with clinical annotations, and prior knowledge enter a structured encoder (Figure 1B). Cellular state $Z_{\mathrm{cell}}$, tissue immune context $Z_{\mathrm{tme}}$, and individual context $Z_{\mathrm{patient}}$ are combined into the structured state $S_t$, while the intervention under evaluation is encoded independently as $A_t$. Treatment information within the individual context denotes baseline or prior treatment history rather than the action being predicted. The state and action jointly parameterize the transition

\begin{equation}
S_{t+1}=F_{\theta}(S_t,A_t),
\end{equation}

where $A_t$ may specify a cytokine, genetic edit, drug, checkpoint intervention, or combination. For bounded counterfactual rollout, a subsequent action can be composed as

\begin{equation}
S_{t+2}=F_{\theta}(S_{t+1},A_{t+1}).
\end{equation}

Structured decoders produce expression-level predictions and interpretable biological readouts. These include cell proportions, program scores, pathway shifts, differential-expression direction, nonresponder-to-responder scores, and mechanism explanations. The outputs support mechanistic interpretation, biomarker discovery, intervention ranking, and hypothesis generation. The architecture supports bounded action composition, whereas the quantitative dynamic benchmarks reported here evaluate the one-step transition to $S_{t+1}$ unless explicitly stated otherwise.

To determine which components were required for this multiscale transition model, we evaluated seven capabilities and performed complementary module- and representation-level ablations (Figures 1C--1E). The capability analysis covered gene-program fidelity, cellular state, tissue context, individual response, action-conditioned dynamics, cross-context transfer, and mechanistic or counterfactual analysis. Comparator models addressed different subsets of these tasks, with scFoundation providing a pretrained single-cell representation reference.\cite{ref13} Omitting the structured decoder produced a 0.14 loss in composite score, and omitting action-conditioned dynamics produced a 0.10 loss. Substitution of the action representation, multiscale encoder, or immune-adapted cell encoder lowered the score by 0.07, 0.08, and 0.06, respectively. At the representation level, the structured orthogonal state reached 0.748 balanced accuracy on the held-out rheumatoid arthritis (RA) out-of-distribution (OOD) task, versus 0.616, 0.586, and 0.521 for progressively less structured alternatives. The gated residual action model reached an unseen-context $\Delta\cos$ score of 0.388, compared with 0.369 for an orthogonal residual, 0.296 for a context mean, and 0.248 for a raw action vector. Together, these ablations identified distinct contributions from state organization, action encoding, and biological decoding.

\begin{figure}[!p]
\centering
\begin{minipage}{\textwidth}
\centering
\includegraphics[width=\textwidth,height=0.70\textheight,keepaspectratio]{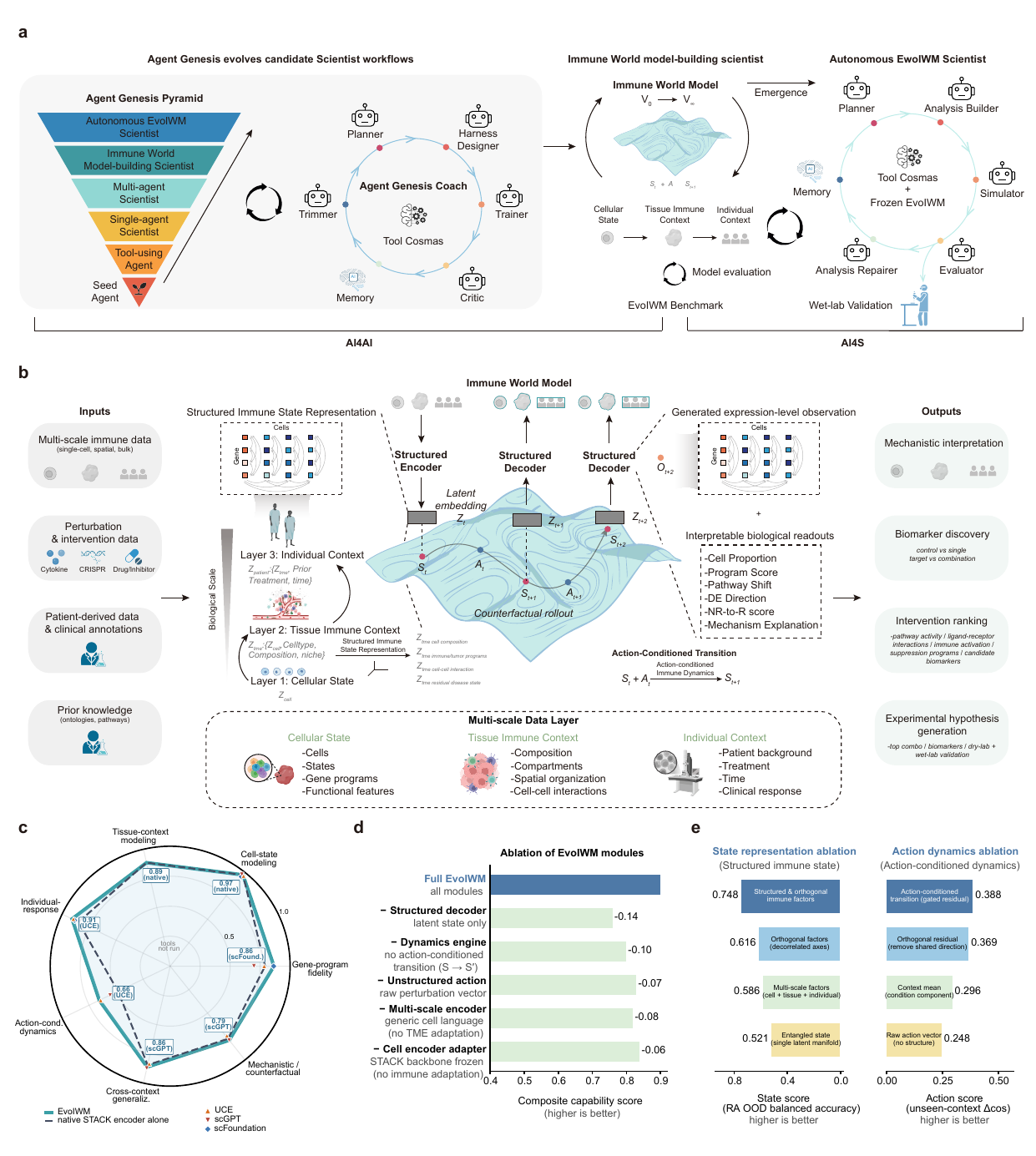}

\vspace{0.4em}
\raggedright\footnotesize
\textbf{Figure 1: Governed construction and component contributions of the Immune World Model.} (A) Agent Genesis evolves candidate Scientist workflows; the retained Immune World Model--building Scientist develops the Immune World Model, and the fixed Immune World Model Research Scientist subsequently operates around the frozen model and prioritizes selected hypotheses for prospective experimental testing. (B) Multiscale evidence is encoded as structured cellular, tissue, and individual states. Explicit actions condition the transition from $S_t$ to $S_{t+1}$ and can be composed for a bounded counterfactual rollout to $S_{t+2}$. Structured decoders return predicted outputs and interpretable biological readouts for mechanism-oriented analysis, candidate biomarker prioritization, intervention ranking, and hypothesis generation. (C) Capability coverage across seven world-model dimensions. (D) Module ablation of the assembled system. (E) State-representation ablation scored by RA OOD balanced accuracy and action-dynamics ablation scored by unseen-context $\Delta\cos$. Higher values indicate better performance.
\end{minipage}
\end{figure}

\subsection*{Governed development selects robust model improvements}

We first tested whether continuous scoring could distinguish candidate builders once binary qualification had saturated. In a six-builder comparison, macro-F1 was 0.681 for the Immune World Model--building Scientist, 0.665 for DeepSeek V4, 0.662 for Claude S5, 0.658 for GPT-5.6, 0.646 for Gemini 3.6, and 0.629 for Qwen 3.6 (Figure 2A). Because the leading systems all passed the G4 qualification boundary, subsequent lineage selection used the continuous development-set metric.

Locked external-capability tests then examined whether the frozen Immune World Model improved over prespecified baselines beyond the builder comparison (Figure 2B). Paired Pearson gains were $+0.112$ for known compounds in held-out cell contexts ($n=151$), $+0.054$ for unseen double-gene combinations ($n=23$), $+0.293$ for unseen drugs in held-out contexts ($n=30$), $+0.159$ for unseen-gene transfer from K562 to RPE1 ($n=200$), and $+0.481$ for unseen-gene transfer from 8 to 48 h ($n=200$). Gains were positive in all five locked probes, although their magnitudes differed substantially across transfer settings.

The resulting Immune World Model lineage comprised the initial U000 configuration and 15 accepted successors, while nine rejected proposals remained recorded without entering the accepted lineage (Figure 2C). Updates to cellular state, tissue context, individual response, action dynamics, cross-context transfer, and mechanistic reasoning occurred at different iterations. Retaining both promoted and rejected proposals placed the final Immune World Model configuration within an auditable development history.

We next distinguished system-level audit from promotion of individual model classes (Figure 2D). The locked cross-system audit compared OOD performance, execution reliability, repair-selection quality, seed stability, and development--test alignment across the fixed Immune World Model Research Scientist and five external systems.

Within the Immune World Model lineage, the study-specific Immune68-to-TabPFN update supplied the frozen Immune68 immune-feature table to the Tabular Prior-data Fitted Network (TabPFN)\cite{ref45} and achieved a paired training leave-one-context-out (LOCO) macro-F1 gain of $+0.243$, clearing the predeclared $+0.02$ promotion gate. Balanced TabPFN, the class-rebalanced candidate, and Wide-state TabPFN, the expanded-state-input candidate, were rolled back after changes of $-0.074$ and $-0.092$, respectively. Thus, a proposed model class entered the lineage only when its locked improvement cleared the gate.

A direct comparison of two candidate Immune World Model updates illustrated how the gate distinguished an update meeting all prespecified criteria from a marginal one (Figure 2E). U019 achieved a Pearson-correlation gain of $+0.293$ (95\% CI, $0.250$--$0.335$) and passed all seven gates. U020 achieved a gain of $+0.013$ (95\% CI, $0.005$--$0.022$) but failed the predeclared $+0.020$ effect-size gate despite passing the other six checks. Statistical positivity alone therefore did not justify promotion.

\begin{figure}[!p]
\centering
\begin{minipage}{\textwidth}
\centering
\includegraphics[width=\textwidth,height=0.70\textheight,keepaspectratio]{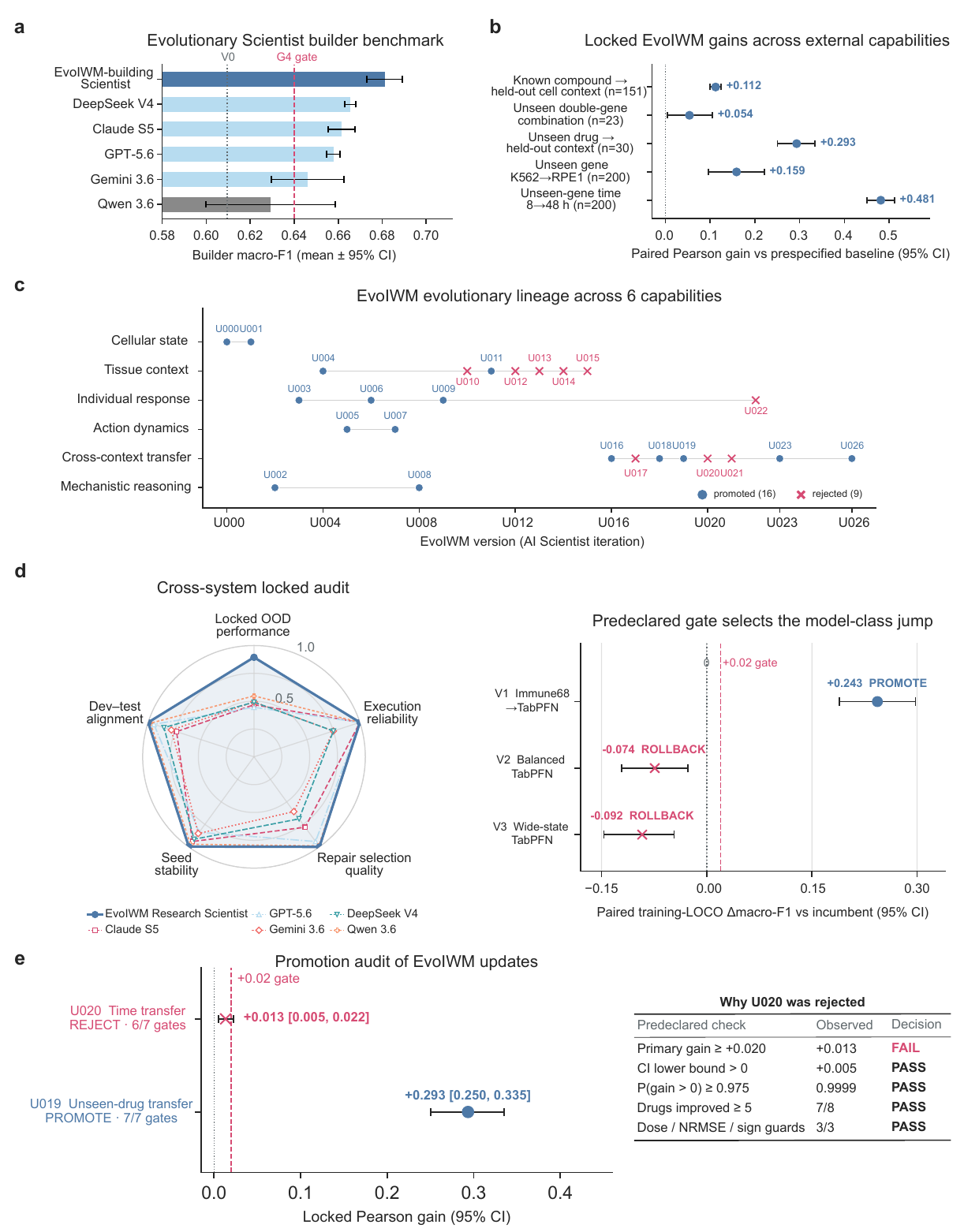}

\vspace{0.4em}
\raggedright\footnotesize
\textbf{Figure 2: Locked evaluation and predeclared gates govern immune-world-model development.} (A) Continuous builder macro-F1 distinguishes candidate development systems; points and intervals show means and 95\% confidence intervals. (B) Locked paired Pearson gains for the frozen Immune World Model across five external-capability tests. (C) Versioned Immune World Model lineage across six capability lanes; 16 promoted and nine rejected candidates remain visible. (D) Cross-system locked OOD audit and gated model-class selection. The Immune68-to-TabPFN update couples the frozen study-specific immune-feature table to TabPFN and is promoted at $+0.243$; class-rebalanced Balanced TabPFN and expanded-input Wide-state TabPFN are rolled back at $-0.074$ and $-0.092$. (E) Immune World Model update U019 passes all seven gates with a gain of $+0.293$ (95\% CI, $0.250$--$0.335$); U020 is rejected despite six passing checks because its $+0.013$ gain (95\% CI, $0.005$--$0.022$) does not reach the predeclared $+0.020$ threshold.
\end{minipage}
\end{figure}

\subsection*{The frozen model supports immune-state forecasting across scales}

With the retained configuration frozen, we next tested whether the Immune World Model could forecast immune-state transitions from cellular perturbation responses through tissue and individual context. The Parse PBMC few-shot task was derived from a cytokine-perturbation atlas of approximately 10 million PBMCs from 12 donors exposed to 90 cytokines;\cite{ref49} the benchmark retained 86 cytokine contexts after preprocessing. Pearson $\Delta$ increased from 0.309 for a cell-mean baseline to 0.427 for an additive model, 0.462 for Rhaister, a summary-statistic perturbation-response predictor,\cite{ref46} and 0.495 for the Immune World Model (Figure 3A), yielding the highest point estimate in the low-data setting.

The displayed program-level coupling remained high after cytokine-family exclusions (Figure 3B). Pearson correlation was 0.98 across all 86 cytokines, 0.98 after excluding interferons ($n=78$), 0.99 after excluding $\gamma_c$ cytokines ($n=80$), and 0.97 after excluding both groups ($n=72$). One-step action-conditioned prediction further distinguished the effects of IL-2, IL-7, and IL-15 across MHC-I, MHC-II, CXCR3, checkpoint, and interferon-stimulated-gene programs (Figure 3C).

In a common gene space, GEARS, the additive baseline, and scGPT reached mean $\Delta\cos$ scores of 0.25, 0.28, and 0.31, whereas the condition-mean model and the Immune World Model both reached 0.46 (Figure 3D); the Immune World Model therefore did not improve the global mean over the condition-mean baseline. Relative to scGPT, however, the Immune World Model achieved a higher cosine similarity for 404 of 636 actions (64\%), with a median per-action gain of $+0.22$.

Tissue context changed both the magnitude and direction of the inferred anti-PD-1 responder axis (Figure 3E). Full-tumor-microenvironment inputs linked response in HCC and head and neck squamous cell carcinoma (HNSCC) to increased cytotoxic and decreased naive/memory programs, whereas melanoma showed the opposite inferred direction and basal cell carcinoma (BCC) showed a smaller reversal. For C1--C6 immune-ecosystem prediction,\cite{ref1} macro-F1 was 0.52 for cellular features, 0.60 for tissue features, 0.64 for their combination, and 0.15 for a shuffled control (Figure 3F), indicating that arbitrary context did not reproduce the tissue contribution.

\begin{figure}[!p]
\centering
\begin{minipage}{\textwidth}
\centering
\includegraphics[height=0.73\textheight,keepaspectratio]{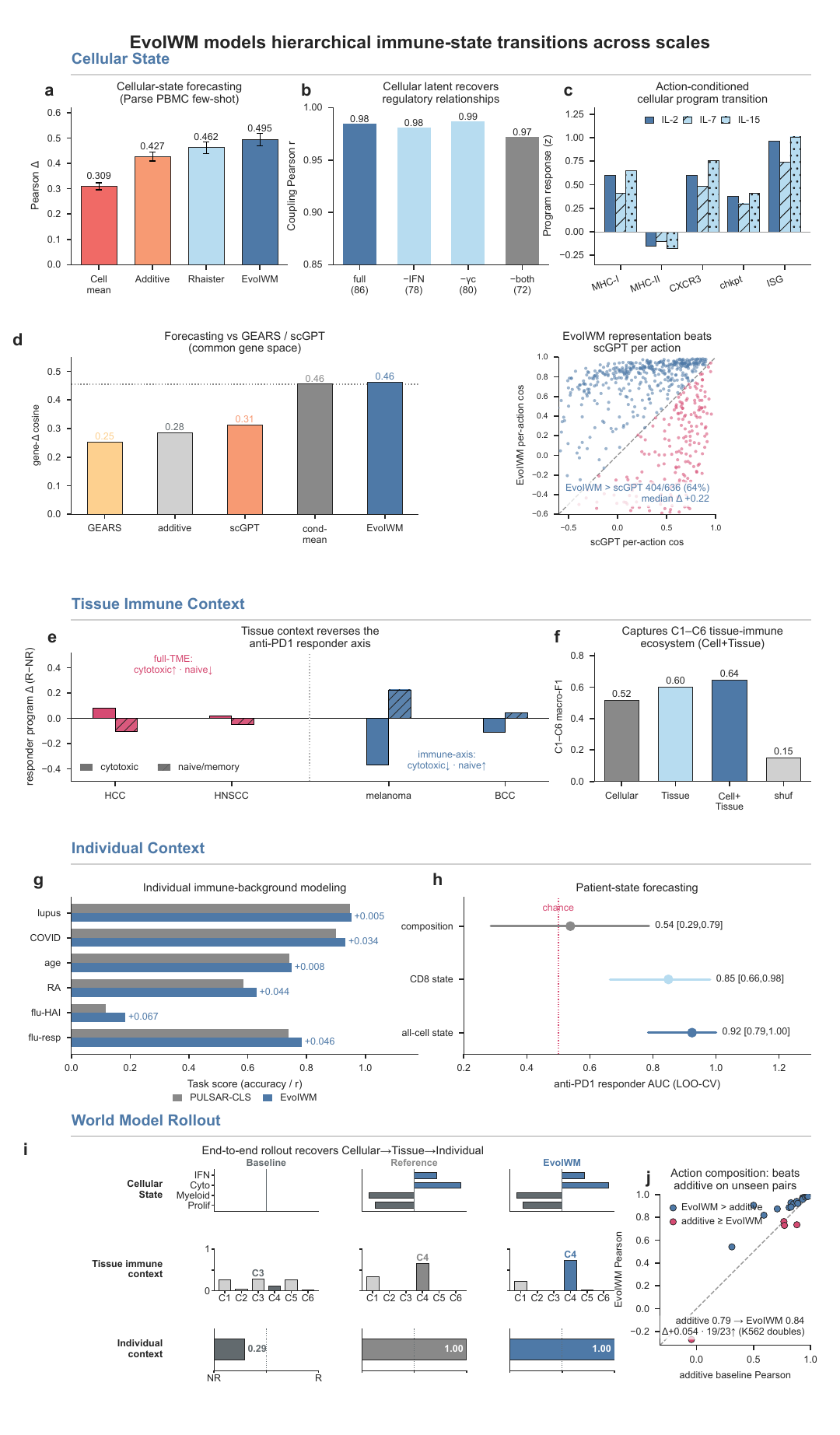}

\vspace{0.4em}
\raggedright\footnotesize
\textbf{Figure 3: The frozen Immune World Model forecasts hierarchical immune-state transitions.} (A) Parse PBMC few-shot cellular-state forecasting scored by Pearson $\Delta$. (B) Regulatory coupling after interferon and $\gamma_c$ cytokine-family exclusions. (C) IL-2-, IL-7-, and IL-15-conditioned transitions across five cellular programs. (D) Gene-$\Delta$ cosine forecasting against GEARS, additive, scGPT, and condition-mean baselines in a common gene space, together with the per-action Immune World Model--scGPT comparison. (E) Tissue-dependent reversal of the anti-PD-1 responder axis. (F) C1--C6 tissue-immune ecosystem macro-F1 using cellular, tissue, combined, and shuffled inputs. (G) Individual immune-background tasks. (H) Anti-PD-1 patient-state forecasting by composition, CD8 state, and all-cell state. (I) Representative end-to-end Cellular$\rightarrow$Tissue$\rightarrow$Individual World Model Rollout. (J) Action composition on unseen pairs compared with an additive baseline.
\end{minipage}
\end{figure}

At the individual scale, the Immune World Model improved performance on all six immune-background tasks relative to the PULSAR donor-representation classification baseline (PULSAR-CLS).\cite{ref47} Gains were $+0.005$ for lupus, $+0.034$ for COVID-19, $+0.008$ for age, $+0.044$ for rheumatoid arthritis, $+0.067$ for influenza hemagglutination-inhibition (HAI) response, and $+0.046$ for influenza response (Figure 3G). Anti-PD-1 leave-one-out AUC was 0.54 (95\% CI, 0.29--0.79) for composition, 0.85 (0.66--0.98) for CD8 state, and 0.92 (0.79--1.00) for all-cell state (Figure 3H), consistent with functional state carrying predictive information beyond cell abundance.

We then examined whether these scale-specific predictions could be connected in an illustrative rollout. In the representative end-to-end example, the baseline favored tissue state C3 and produced an individual responder score of 0.29, whereas the reference and the Immune World Model both selected C4 and reached 1.00 at the individual level (Figure 3I).

Action composition also improved prediction on unseen pairs: Pearson correlation increased from 0.79 for the additive baseline to 0.84 for the Immune World Model, a gain of $+0.054$, with improvement for 19 of 23 pairs and particularly large gains for K562 double perturbations (Figure 3J).

\subsection*{Complementary immune axes nominate IL-36$\gamma$ plus SIRP$\alpha$ inhibition}

To move from forecasting to hypothesis generation, we combined measured perturbations with model inference to identify complementary immune axes as candidate therapeutic hypotheses. In the surface-knockout screen, CD47 was among the strongest druggable shifts toward a hot or inflamed state (Figure 4A). CD47 is an established self-recognition signal whose blockade can promote tumor-cell phagocytosis and link myeloid uptake to adaptive antitumor responses.\cite{ref38,ref39,ref40,ref41,ref42} CD47 knockout supplied evidence for the innate-checkpoint arm; because SIRP$\alpha$ is the cognate inhibitory receptor of CD47, SIRP$\alpha$ inhibition served as the therapeutic counterpart of that perturbation.

The measured cytokine screen ranked IFN-$\beta$, IL-15, IFN-$\omega$, IFN-$\gamma$, and IL-2 highest (Figure 4B). Zero-shot upstream nomination of the IL-36$\gamma$/IL-18 route exceeded an additive reference under OrthJEPA, VICReg,\cite{ref48} and SE-PCA. OrthJEPA was the study-specific orthogonal embedding, VICReg the published regularization objective, and SE-PCA a principal-component reference fitted to standardized expression. The nomination was recovered in 86 of 86, 80 of 86, and 67 of 86 contexts. Within the locked ranking, IL-36$\gamma$ was the lead upstream candidate, while IL-18 provided secondary support for the broader cytokine-response axis. Prior studies link IL-36$\gamma$ to type-1 lymphocyte programs and identify IL-18-binding protein as a regulator of IL-18 immunotherapy, providing biological context for this axis.\cite{ref43,ref44}

The knockout-by-cytokine analysis showed different modeled interactions for the two arms (Figure 4C). CD47 showed a comparatively weak interaction of $-0.18$, versus $-0.55$ for STAT1, $-0.79$ for JAK1, $-0.80$ for JAK2, $-0.88$ for IFNGR1, and $-1.12$ for MYC. By the displayed criterion, CD47 was comparatively weakly coupled to the interferon-cytokine axis. Pan-cancer expression and axis scores for \textit{CD47}, \textit{SIRPA}, and \textit{IL18} varied across ten tumor types, defining tumor contexts in which both axes were represented (Figure 4D).

Compartment deletion distinguished model dependence from standalone predictive performance (Figure 4E). CD8 removal caused a 0.48 loss, Treg removal caused a 0.02 loss, and removal of the other displayed lineages produced no measurable loss. Standalone AUCs were 0.87 for B cells, 0.85 for CD8 T cells, 0.81 for CD4 T cells, 0.71 for Treg and myeloid states, 0.49 for dendritic cells, and 0.48 for NK cells. The displayed responder-state transition showed the largest model dependence on the CD8 program, whereas several other compartments carried standalone predictive signal.

No cytokine pair passed all prespecified filters. Sequential filtering reduced the 276 cytokine pairs to 193 above the initial $>0.10$ threshold, 77 with net induction, five that exceeded the best single agent, five that remained robust across eight seeds, three that exceeded the raw baseline, one supported by literature adjudication, and none that passed the dominance audit (Figure 4F). This negative audit redirected nomination toward the measured CD47--SIRP$\alpha$ arm and the inferred IL-36$\gamma$/IL-18--IFN arm as complementary axes for prospective testing.

Taken together, these analyses supported an innate checkpoint axis and an adaptive axis linked to IFN. The locked cytokine ranking placed IL-36$\gamma$ ahead of IL-18, and IL-36$\gamma$ plus SIRP$\alpha$ inhibition was selected as the prioritized combination representing these axes. This nomination is a model-derived hypothesis; its experimental evaluation lies outside the scope of the present study.

\begin{figure}[!p]
\centering
\begin{minipage}{\textwidth}
\centering
\includegraphics[height=0.72\textheight,keepaspectratio]{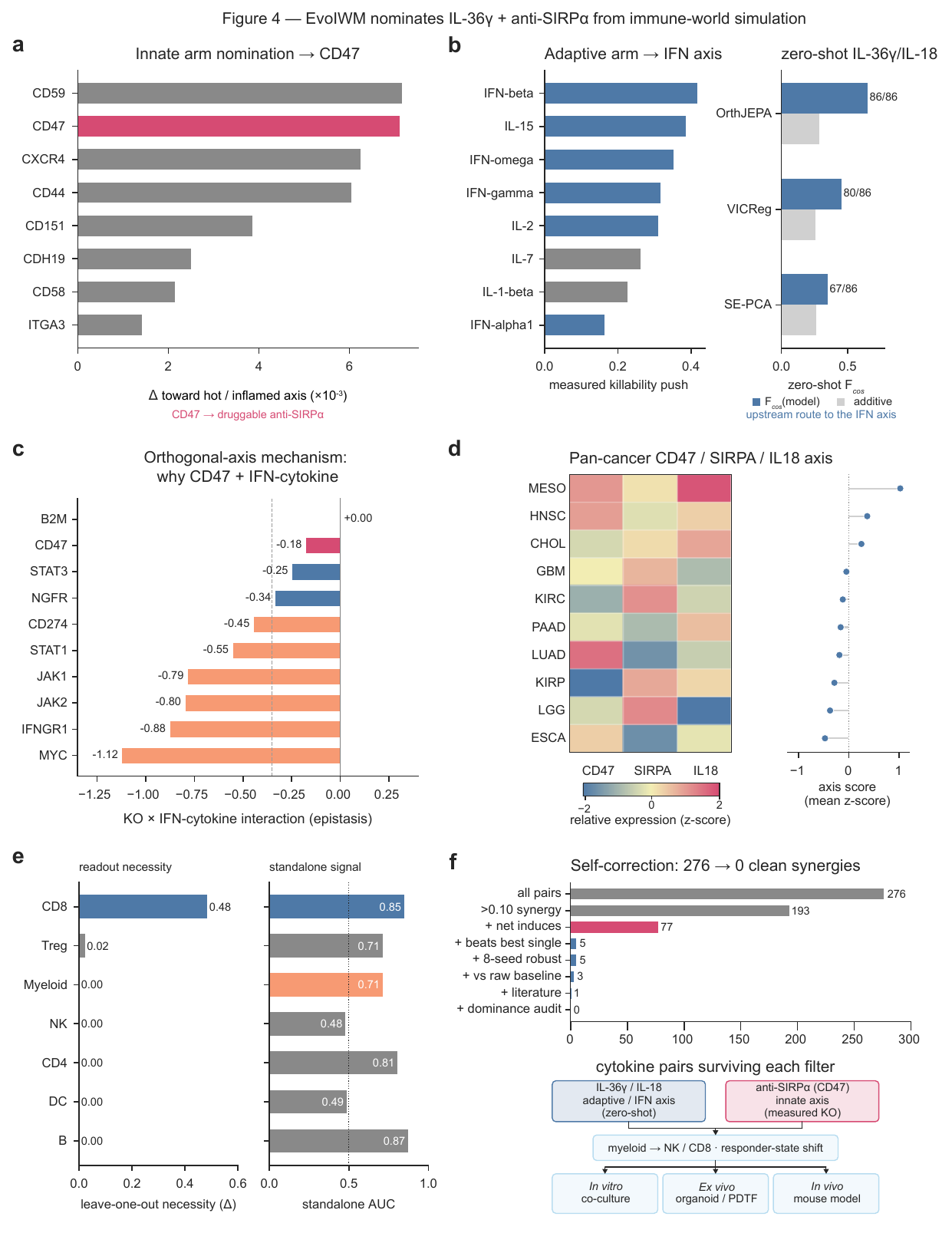}

\vspace{0.4em}
\raggedright\footnotesize
\textbf{Figure 4: Complementary-axis evidence nominates IL-36$\gamma$ plus SIRP$\alpha$ inhibition.} (A) A measured surface-knockout screen places CD47 among the strongest druggable shifts toward a hot or inflamed state. (B) A measured cytokine screen identifies an interferon-associated killability axis. Zero-shot IL-36$\gamma$/IL-18 nomination exceeds the additive reference under OrthJEPA (86/86 contexts), VICReg (80/86), and SE-PCA (67/86). (C) Knockout-by-interferon-cytokine epistasis compares CD47 with canonical pathway components. (D) Pan-cancer \textit{CD47}, \textit{SIRPA}, and \textit{IL18} expression and mean axis scores across ten tumor types characterize expression context. (E) Leave-one-lineage-out model dependence is compared with standalone lineage AUC. (F) Sequential filtering of 276 screened cytokine pairs, none of which passed all filters. Together, the panels identify two complementary immune axes and nominate IL-36$\gamma$ plus SIRP$\alpha$ inhibition as a prospectively testable combination.
\end{minipage}
\end{figure}

\clearpage
\section*{DISCUSSION}

The Immune World Model reframes immune prediction as the estimation of an intervention-conditioned state transition rather than the assignment of a static sample label. Single-cell foundation and perturbation models have made transferable cell-state encoding and extrapolation increasingly tractable, but therapeutic effects remain conditioned by tissue ecology and host immune context. By representing the intervention independently from cellular, tissue, and individual state, the Immune World Model estimates how a specified action moves a biological system through immune-state space. The benefit of combining cell and tissue information, together with the higher anti-PD-1 AUC point estimate obtained from functional all-cell states than from composition alone, supports the value of structured multiscale context. World Model Rollout extends the same formulation across biological scales, providing a common interface between molecular programs, tissue ecosystems, and patient-level readouts.

The Immune World Model--building Scientist constructed the model while addressing a distinct problem: how to develop such a model while preserving a testable record of the choices that produced it. Agent Genesis treated architecture changes, repairs, and evaluation procedures as versioned proposals evaluated under common budgets and prespecified gates. Consequently, both successful promotions and rejected alternatives remained part of the evidence for the retained world model. This is important because adaptive model development can otherwise blur the distinction between improvement of a biological predictor and repeated accommodation to an evaluation set. The present results show that an AI Scientist can construct a predictive model not only by generating candidates, but also by organizing comparison, rollback, and model freezing before confirmation. Agent Genesis is therefore the governed model-construction system, whereas the Immune World Model is the resulting biological transition model, evaluated here on locked computational benchmarks. After freezing, the fixed Immune World Model Research Scientist used the model without modifying its parameters or configuration.

Across the confirmation tasks, the most informative evidence came from convergence among different forms of generalization. The Immune World Model preserved cytokine-regulatory relationships after cytokine-family exclusions, distinguished program-level responses to related cytokines, incorporated tissue context into immune-ecosystem prediction, and improved 19 of 23 unseen perturbation pairs. The tie between the Immune World Model and the condition-mean baseline in global mean gene-$\Delta$ cosine indicates that aggregate performance does not fully capture action-level improvements, withheld-context behavior, or cross-scale propagation of structured readouts. Together, these results are consistent with a shared transition representation that captures context-dependent structure beyond a raw action vector or an averaged perturbation effect.

The therapeutic analysis illustrates how this representation can be used to formulate a biological hypothesis. The measured CD47 knockout phenotype nominated an innate checkpoint arm, whereas cytokine-response modeling prioritized an IL-36$\gamma$-associated adaptive/IFN arm. The comparatively weak modeled interaction between the arms suggested that they could provide nonredundant immune inputs. One mechanistic interpretation is that relieving CD47--SIRP$\alpha$ restraint may facilitate myeloid uptake and antigen handling, while IL-36$\gamma$ reinforces type-1 lymphocyte programs and CD8 T-cell and NK-cell effector activity.\cite{ref38,ref39,ref40,ref41,ref42,ref43} The combination is therefore presented as a model-derived, prospectively testable hypothesis rather than a validated therapy. Experimental confirmation, and any claim of formal pharmacological synergy, remains to be established.

The current study also defines priorities for further development. These include cohort-level validation of recursive rollout, clinical-response analysis in larger external cohorts, and prospective experimental evaluation of the nominated combination with dose-response, target engagement, pharmacodynamic, and toxicity assessments. Incorporating spatial organization, surface-protein state, receptor occupancy, clonality, treatment exposure, longitudinal sampling, and calibrated uncertainty should further improve biological resolution and forecast prioritization. With these extensions, governed immune world models could provide a general framework for connecting multiscale perturbation data to prospectively testable therapeutic hypotheses.

\section*{RESOURCE AVAILABILITY}

\subsection*{Lead contact}

Questions about study resources should be sent to the lead contact, Yingcheng Wu (\href{mailto:wuyc@phai-labs.com}{wuyc@\allowbreak phai-labs.com}).

\subsection*{Materials availability}

This study is computational and did not generate new reagents or materials.

\subsection*{Data and code availability}

\begin{itemize}[leftmargin=*]
\item Public dataset accessions are provided by the cited source publications. Inclusion, quality-control, and partitioning records will be released in an accompanying versioned manifest.
\item A versioned software release will contain Agent Genesis and Immune World Model--building Scientist selection, transition-model fitting, evaluation gates, one-step forecasts, bounded counterfactual rollout, World Model Rollout analyses, and figure-generation code.
\item Subject to data-use agreements, consent, and privacy constraints, the release will include run packets, lineage decisions, and sample-level model outputs.
\end{itemize}

\section*{DECLARATION OF INTERESTS}

The authors declare no competing interests.

\section*{DECLARATION OF GENERATIVE AI AND AI-ASSISTED TECHNOLOGIES}

Codex assisted with LaTeX assembly, prose editing, and consistency checks. The authors reviewed and revised all AI-assisted text and take responsibility for the submitted manuscript.

\begingroup
\small
\bibliographystyle{numbered}
\bibliography{references}
\endgroup

\clearpage
\section*{STAR METHODS}

\subsection*{Data and study participant details}

\subsubsection*{Public immune and cancer cohorts}

Cellular-state and action-conditioned forecasting used the Parse PBMC cytokine atlas, which comprises 9,697,974 post-quality-control PBMC profiles from 12 donors (six female and six male; ages 34--75 years) exposed for 24 h to one of 90 cytokines or PBS.\cite{ref49} Donor, cytokine, dose, and one of 16 retained cell-type states defined each matched transition. The locked benchmark retained 86 cytokine contexts after preprocessing, yielding 14,577 donor-matched control--action transitions. Common-gene-space perturbation benchmarks supported comparisons with GEARS and scGPT. Tumor and anti-PD-1 cohorts supported tissue-context, C1--C6 immune-ecosystem, and patient-state analyses. Donor-background datasets spanning lupus, COVID-19, age, rheumatoid arthritis, influenza HAI response, and influenza response supported individual-context prediction. Surface-knockout screens, cytokine-interaction data, and pan-cancer expression profiles supported nomination of therapeutic hypotheses. All analyses used previously published or publicly deposited human data, and this study did not generate new experimental data.

\subsection*{Method details}

\subsubsection*{Versioned Evolutionary Scientist Engine}

The Evolutionary Scientist Engine comprised an AI4AI architecture-evolution stage and an AI4S research stage. In AI4AI, the Agent Genesis Pyramid organized the following successive candidate classes: seed agent, tool-using agent, single-agent Scientist, multi-agent Scientist, Immune World Model--building Scientist, and Immune World Model Research Scientist. The Agent Genesis Coach coordinated a planner, harness designer, trainer, critic, memory, trimmer, and Tool Cosmas. The planner proposed architectural or workflow changes; the harness designer instantiated comparable execution and evaluation environments; the trainer fitted trainable components; the critic diagnosed failures; memory retained prior proposals and evidence; the trimmer removed unnecessary components; and Tool Cosmas controlled access to approved analytical tools. In AI4S, the fixed Immune World Model Research Scientist coordinated a planner, builder, simulator, evaluator, repairer, memory, and Tool Cosmas around the frozen model, benchmarked model behavior, and prioritized selected hypotheses for prospective experimental evaluation. The Immune World Model Research Scientist operated within the versioned research harness without modifying the frozen model.

Candidate Scientist and model changes were assigned immutable version identifiers. Development and confirmation data were separated by immutable manifests before lineage evolution. Proposals, repairs, and promotion decisions used the development manifest and its prespecified metric. The retained Immune World Model Research Scientist workflow and Immune World Model configuration were frozen before the independent confirmation manifest was opened, and confirmation was performed without further adaptation. Accepted changes updated the lineage frontier, while all promotion decisions and associated evidence were retained. Hypothesis prioritization and biological interpretation were conducted under investigator oversight.

\subsubsection*{Builder benchmark}

Candidate builders were compared using continuous builder macro-F1. Figure 2A reports means and 95\% confidence intervals for the Immune World Model--building Scientist, DeepSeek V4, Claude S5, GPT-5.6, Gemini 3.6, and Qwen 3.6 and shows a G4 qualification boundary.

\subsubsection*{Locked external-capability and gate audit}

The five Figure 2B probes were evaluated against prespecified baselines using paired Pearson gain and locked manifests: known compounds in held-out cell contexts, unseen double-gene combinations, unseen drugs in held-out contexts, unseen-gene transfer from K562 to RPE1, and unseen-gene transfer from 8 to 48 h. The cross-system audit of the fixed Immune World Model Research Scientist in Figure 2D summarized locked OOD performance, execution reliability, repair-selection quality, seed stability, and development--test alignment. Model-class updates were compared with the incumbent by paired training leave-one-context-out (LOCO) macro-F1, in which each biological context was excluded in turn and candidate-minus-incumbent macro-F1 was computed on the excluded context. Immune68 denotes the frozen study-specific immune-feature table; the Immune68-to-TabPFN candidate supplied this table to TabPFN.\cite{ref45} Balanced TabPFN used the class-rebalanced training configuration, whereas Wide-state TabPFN used the expanded state input. These labels denote internal candidates rather than separately published algorithms. The primary promotion threshold was fixed at $+0.020$. Figure 2E additionally checked confidence-interval positivity, posterior probability of gain, and the number of improved drugs, together with dose, normalized root-mean-square error (NRMSE), and sign guards. The dose guard required preservation of the prespecified dose-response ordering, the NRMSE guard prohibited an increase beyond its locked tolerance, and the sign guard required preservation of the expected direction of the perturbation effect.

\subsubsection*{Hierarchical immune-state representation}

The structured encoder integrated four evidence classes shown in Figure 1B: multiscale immune measurements, perturbation and intervention data, patient-derived data with clinical annotations, and prior knowledge such as ontologies and pathways. Layer 1 encoded cellular state as $Z_{\mathrm{cell}}$, including cell identity, state, gene programs, and functional features. Layer 2 encoded tissue immune context as $Z_{\mathrm{tme}}$, combining cellular representations with cell type, composition, spatial niche, and cell--cell interaction information. Layer 3 encoded individual context as $Z_{\mathrm{patient}}$, combining tissue context with patient background, prior treatment history, time, and clinical response. These layers defined the structured latent state $S_t$. State-representation variants compared structured orthogonal immune factors, orthogonal factors without the full structure, non-orthogonal multiscale factors, and an entangled latent state.

\subsubsection*{Action-conditioned immune dynamics}

Actions $A_t$ represented cytokines, genetic perturbations, drugs, immunotherapies, or combinations and were encoded separately from the biological context in $S_t$. The core transition predicted $S_{t+1}=F_{\theta}(S_t,A_t)$. For bounded counterfactual rollout, a subsequent action was composed as $S_{t+2}=F_{\theta}(S_{t+1},A_{t+1})$, allowing comparison of control, single-target, and combination trajectories. Figure 3 benchmarks evaluate the one-step cellular transition to $S_{t+1}$, while the composed transition represents the bounded two-action formulation. Variants included gated residual action conditioning, an orthogonal residual, a context-mean predictor, and a raw action vector.

\subsubsection*{Structured biological decoder}

Structured decoders mapped predicted states to expression-level predictions and interpretable biological readouts. The displayed readouts included cell proportion, program score, pathway shift, differential-expression direction, nonresponder-to-responder score, and mechanism explanation. These readouts supported mechanistic interpretation, biomarker discovery, intervention ranking, and experimental-hypothesis generation.

\subsubsection*{Capability and ablation analyses}

The composite capability analysis summarized gene-program fidelity, cell-state modeling, tissue-context modeling, individual response, action-conditioned dynamics, cross-context generalization, and mechanistic or counterfactual reasoning. Figure 1C included scFoundation and other foundation-model comparators as task-specific representation references; axes without reported results were left blank. Module ablations removed the structured decoder, action-conditioned dynamics, structured action representation, multiscale encoder, or immune adapter. The state-representation ablation in Figure 1E used held-out RA out-of-distribution balanced accuracy. The action-dynamics ablation used the archived unseen-context $\Delta\cos$ score for action-induced latent displacement.

\subsubsection*{Cellular forecasting and representation benchmarks}

The Parse PBMC few-shot benchmark compared a cell-mean baseline, an additive model, Rhaister,\cite{ref46} and the Immune World Model using the displayed Pearson $\Delta$ score (Figure 3A). ``Few-shot'' denotes fitting with the locked reduced set of observed action--context examples before evaluation on the held-out examples; the exact shot count and sampling seed are recorded in the benchmark manifest. Regulatory coupling was recomputed on the full cytokine set and after exclusion of interferon, $\gamma_c$, or both families (Figure 3B). Action-conditioned cellular programs were evaluated for IL-2, IL-7, and IL-15 across MHC-I, MHC-II, CXCR3, checkpoint, and interferon-stimulated-gene readouts (Figure 3C). Common-gene-space forecasting compared GEARS, an additive model, scGPT, a condition-mean baseline, and the Immune World Model by gene-$\Delta$ cosine, followed by a per-action cosine comparison between the Immune World Model and scGPT (Figure 3D).

\subsubsection*{Tissue and individual-context benchmarks}

Tissue-context analysis compared cytotoxic and naive/memory responder programs across HCC, HNSCC, melanoma, and BCC (Figure 3E). C1--C6 immune-ecosystem prediction used cellular features, tissue features, their combination, and a shuffled-context control (Figure 3F). Individual immune-background modeling was evaluated on lupus, COVID-19, age, rheumatoid arthritis, influenza HAI response, and influenza response relative to PULSAR-CLS, the study-specific classification evaluation of the published PULSAR donor representation.\cite{ref47} Anti-PD-1 patient-state forecasting compared composition, CD8 state, and all-cell state by leave-one-out cross-validated area under the receiver-operating-characteristic curve with the displayed confidence intervals (Figure 3H).

\subsubsection*{World Model Rollout and action composition}

The representative World Model Rollout in Figure 3I propagated a cellular program representation to a C1--C6 tissue state and then to an individual responder score, with baseline, reference, and Immune World Model trajectories shown separately. This cross-scale operation is distinct from bounded counterfactual rollout in Figure 1B. Action composition in Figure 3J compared the Immune World Model with an additive baseline on 23 unseen perturbation pairs using Pearson correlation and the direction of pair-level improvement.

\subsubsection*{Governed self-correction audit}

The cytokine-pair audit applied sequential gates for a $>0.10$ apparent-synergy score, positive net induction relative to the untreated control, improvement over the higher-scoring single agent, replication under the locked rule across eight seeds, improvement over the raw-expression baseline, literature adjudication for known or contradictory mechanisms, and a final dominance audit for effects attributable to one component. The displayed survivor counts were 276, 193, 77, 5, 5, 3, 1, and 0. Pairs were retained when they passed each applicable gate.

\subsubsection*{Therapeutic-hypothesis nomination}

The innate arm was derived from a surface-knockout screen for positive displacement toward a prespecified hot or inflamed immune-state reference and was subsequently filtered for druggability. If $h(\mathbf{z})$ denotes the fixed projection of immune state $\mathbf{z}$ onto that reference axis, the displayed knockout displacement was $\Delta h_k=h(\mathbf{z}_k)-h(\mathbf{z}_{0})$. The adaptive/IFN-associated arm combined a measured cytokine killability screen with zero-shot IL-36$\gamma$/IL-18 nomination. OrthJEPA denotes the study-specific orthogonality-constrained joint-embedding representation, VICReg denotes the published variance--invariance--covariance regularization objective,\cite{ref48} and SE-PCA denotes the study-specific principal-component reference fitted to the standardized expression input. OrthJEPA and SE-PCA are internal configuration labels, not independent published model classes. For each representation, the zero-shot forecast cosine measured alignment between predicted upstream-cytokine displacement $\widehat{\mathbf{u}}$ and the measured interferon-axis displacement $\mathbf{v}_{\mathrm{IFN}}$,

\[
F_{\cos}=\frac{\widehat{\mathbf{u}}^{\mathsf T}\mathbf{v}_{\mathrm{IFN}}}
{\lVert\widehat{\mathbf{u}}\rVert_2\lVert\mathbf{v}_{\mathrm{IFN}}\rVert_2}.
\]

The recovery count is the number of locked contexts, out of 86, in which $F_{\cos}$ exceeded the additive reference. The locked nomination designated IL-36$\gamma$ as the lead upstream candidate and IL-18 as secondary support for the broader cytokine-response axis. The nomination defines a prospectively testable hypothesis and was not evaluated experimentally in this study.

Orthogonality was evaluated using knockout-by-interferon-cytokine epistasis. For knockout $k$ and cytokine condition $c$, the interaction was calculated on the common response scale as

\[
\epsilon_{k,c}=R_{k,c}-R_{k,0}-R_{0,c}+R_{0,0},
\]

where values near zero indicate an approximately additive or independent interaction. More negative values indicate greater cancellation of the cytokine response by the knockout. Expression context was assessed from pan-cancer \textit{CD47}, \textit{SIRPA}, and \textit{IL18} values standardized within each gene as $z=(x-\bar{x})/s_x$ across the displayed tumor types; the displayed axis score was the mean of its constituent standardized values. Compartment attribution compared leave-one-lineage-out necessity with standalone lineage AUC to distinguish readout necessity from predictive information.

\subsection*{Quantification and statistical analysis}

For a multiclass task with classes $1,\ldots,K$, macro-F1 was the unweighted mean $K^{-1}\sum_{k=1}^{K}F1_k$, and balanced accuracy was the unweighted mean of class-specific recalls. The paired training-LOCO $\Delta$ macro-F1 for context $i$ was $F1^{\mathrm{candidate}}_i-F1^{\mathrm{incumbent}}_i$ after both models were trained without that context. The reported point estimate and confidence interval summarize the paired context-level differences rather than unpaired model scores.

For an observed baseline expression vector $\mathbf{x}_{i,0}$, observed action-conditioned vector $\mathbf{x}_{i,a}$, and prediction $\widehat{\mathbf{x}}_{i,a}$, the observed and predicted action-induced displacements were

\[
\mathbf{d}_{i,a}=\mathbf{x}_{i,a}-\mathbf{x}_{i,0},\qquad
\widehat{\mathbf{d}}_{i,a}=\widehat{\mathbf{x}}_{i,a}-\mathbf{x}_{i,0}.
\]

Pearson $\Delta$ was $\operatorname{corr}(\widehat{\mathbf{d}}_{i,a},\mathbf{d}_{i,a})$ over the locked feature set. Gene-$\Delta$ cosine, also denoted $\Delta\cos$ in the architecture ablation, was

\[
\Delta\cos_{i,a}=\frac{\widehat{\mathbf{d}}_{i,a}^{\mathsf T}\mathbf{d}_{i,a}}
{\lVert\widehat{\mathbf{d}}_{i,a}\rVert_2\lVert\mathbf{d}_{i,a}\rVert_2},
\]

and the displayed aggregate was the arithmetic mean across the locked action--context units unless a median was explicitly reported. Paired Pearson gain for unit $i$ was $r_i^{\mathrm{candidate}}-r_i^{\mathrm{baseline}}$. NRMSE was $\sqrt{n^{-1}\sum_{j=1}^{n}(\widehat y_j-y_j)^2}/s_y$, where the target normalization scale $s_y$ and the allowed tolerance were fixed in the locked evaluation manifest before scoring.

Area under the receiver-operating-characteristic curve (AUC or AUROC) used ranked continuous scores; leave-one-out cross-validation (LOO-CV) held out one donor or patient. For lineage $\ell$, necessity was $M_{\mathrm{full}}-M_{-\ell}$, where $M$ is the prespecified responder-state transport score; positive values indicate readout loss after removal. Standalone lineage AUC was computed using only lineage $\ell$ and did not measure necessity or causal effects.

All reported results are computational. The statistical unit for each comparison is the locked action--context pair, biological context, donor, or patient defined by the corresponding evaluation manifest, and no unit is reused within a paired comparison.

\end{document}